\documentclass[11pt]{article}

\usepackage[final]{acl}

\usepackage{tempora}
\usepackage{latexsym}

\usepackage[T1,T2A]{fontenc}

\usepackage[utf8]{inputenc}
\usepackage[main=english,russian]{babel}
\usepackage{microtype}

\usepackage{inconsolata}

\usepackage{graphicx}
\usepackage{url}

\title{Evaluating OpenAI GPT Models for Translation of Endangered Uralic Languages: A Comparison of Reasoning and Non-Reasoning Architectures}

\author{
  Yehor Tereshchenko and Mika Hämäläinen \\
  Metropolia University of Applied Sciences \\
  Helsinki, Finland \\
  \texttt{firstname.lastname@metropolia.fi} \\
  \And
  Svitlana Myroniuk \\
  University of Helsinki \\
  Helsinki, Finland \\
  \texttt{firstname.lastname@helsinki.fi} \\
}

\begin{document}
\maketitle
\begin{abstract}
The evaluation of Large Language Models (LLMs) for translation tasks has primarily focused on high-resource languages, leaving a significant gap in understanding their performance on low-resource and endangered languages. This study presents a comprehensive comparison of OpenAI's GPT models, specifically examining the differences between reasoning and non-reasoning architectures for translating between Finnish and four low-resource Uralic languages: Komi-Zyrian, Moksha, Erzya, and Udmurt. Using a parallel corpus of literary texts, we evaluate model willingness to attempt translation through refusal rate analysis across different model architectures. Our findings reveal significant performance variations between reasoning and non-reasoning models, with reasoning models showing 16 percentage points lower refusal rates. The results provide valuable insights for researchers and practitioners working with Uralic languages and contribute to the broader understanding of reasoning model capabilities for endangered language preservation.
\end{abstract}

\section{Introduction}

The rapid advancement of Large Language Models (LLMs) has revolutionized machine translation \cite{Xu2023Paradigm}, yet their performance on endangered language MT tasks remains largely unexplored. While recent translation systems excel for high-resource language pairs (see \citealp{robinson2023chatgpt}), the challenges of morphological complexity, limited training data and cultural specificity\footnote{For background on Erzya sociolinguistic distribution and domains of use, see \citet{Rueter2013Erzya}.} present unique obstacles for Uralic languages.

The Uralic language family, comprising over 30 languages with varying degrees of endangerment, represents an ideal testbed for evaluating LLM translation capabilities \cite{Pirinen2015Preface}. Languages such as Komi-Zyrian, Moksha, Erzya, and Udmurt face significant challenges in digital representation and computational processing, making them particularly vulnerable to language loss while simultaneously offering rich linguistic diversity for research \cite{Alnajjar2023}.

This study addresses a critical gap in LLM evaluation by conducting a systematic comparison of OpenAI's GPT models, specifically examining the differences between reasoning and non-reasoning architectures for translating from Finnish to four low-resource Uralic languages. Our research questions focus on: (1) How do reasoning models compare to non-reasoning models for Uralic language translation willingness? (2) What are the performance differences between model sizes and architectures in terms of refusal rates? (3) Which Uralic languages present the greatest challenges for different model types?

Our contributions include: (1) the first comprehensive evaluation of reasoning vs non-reasoning models for Uralic language translation willingness, (2) a systematic comparison of different GPT architectures using refusal rate analysis, (3) identification of language-specific challenges across Uralic languages, and (4) practical insights demonstrating superior performance of reasoning models for low-resource language tasks.

Beyond methodological interest, refusal behavior also raises ethical concerns for language equity and access; recent work has analyzed ethical and safety gaps in LLMs \citep{tereshchenko2025comparativeanalysisethicalsafety}.
Moreover, LLM behavior in domain- and resource-constrained contexts can complicate downstream NLP pipelines and embeddings—for instance, detecting policy-violating content in fast, noisy gaming chats benefits from tailored embeddings and fine-tuned transformers over generic LLM prompting \citep{Tereshchenko2025ToxicityGaming}.

\section{Related Work}

\subsection{Machine Translation for Low-Resource Languages}
The challenge of machine translation for low-resource languages has been a persistent focus in computational linguistics. Traditional approaches have relied heavily on statistical machine translation (SMT) methods, which require substantial parallel corpora for effective training \citep{Koehn2007}. The advent of neural machine translation (NMT) brought new possibilities through sequence-to-sequence models, yet the fundamental challenge of limited training data remained \citep{Bahdanau2014}.

Recent advances in multilingual NMT have shown promise for low-resource languages through transfer learning and zero-shot translation capabilities \citep{Johnson2017}. However, these approaches still require significant amounts of monolingual data and may not adequately capture the linguistic diversity of endangered languages, which has been tried to tackle with rule-based generation \cite{alnajjar-etal-2023-bootstrapping}. The emergence of large language models has introduced new paradigms for translation that do not require task-specific training, potentially offering solutions for languages with minimal digital resources.

\subsection{Large Language Models for Translation}
Large Language Models have demonstrated remarkable capabilities in translation tasks across various language pairs, often outperforming specialized translation systems \citep{Hendrycks2021}. The zero-shot and few-shot capabilities of models like GPT-3 and GPT-4 have shown particular promise for low-resource language scenarios \citep{Brown2020}.

Recent studies have explored the translation capabilities of LLMs across different language families, revealing both strengths and limitations. While these models excel at high-resource language pairs, their performance on morphologically complex and low-resource languages remains understudied. The few-shot learning paradigm has shown particular promise for adapting to new languages with minimal examples \citep{Wei2022}.

However, systematic evaluation of LLMs for endangered and low-resource languages has been limited. Most studies focus on major world languages, leaving a significant gap in understanding how these models perform on languages with limited digital presence and complex morphological structures.

\subsection{Uralic Language Processing}
The Uralic language family presents unique challenges for computational linguistics due to its agglutinative morphology and complex case systems. Recent work has focused on developing computational resources for Uralic languages, including morphological analyzers \cite{rueter-etal-2020-open}, syntactic parsers and machine translation systems \citep{Tyers2019}.

The computational processing of Uralic languages has gained increasing attention, particularly for languages like Finnish, Estonian, and Hungarian, which have more substantial digital resources \cite{Proszeky2011Endangered,hamalainen2021current}. However, many Uralic languages, face significant challenges in digital representation and computational processing \citep{Partanen2018,hamalainen2021documentacion}.

Recent advances in multilingual language models have shown promise for Uralic languages, with particular success in morphological analysis \cite{hamalainen2021neural} and syntactic parsing \citep{Voutilainen2019}. However, machine translation for Uralic languages remains challenging due to the complex morphological structures and limited parallel corpora available for training.

The unique agglutinative nature of Uralic languages presents specific challenges for computational processing, particularly in machine translation where morphological complexity can lead to significant translation errors. Recent work has explored the use of linguistic knowledge in improving translation quality for Uralic languages, with mixed results \citep{Partanen2020}. Additionally, sentiment analysis research has demonstrated the effectiveness of aligned word embeddings for Uralic languages \citep{Alnajjar2023}, providing insights into cross-lingual representation learning that may inform translation approaches.

\section{Methodology}

This section describes our experimental methodology, including the dataset, model selection, evaluation metrics, and experimental setup.

\subsection{Dataset}
We utilize a parallel corpus consisting of literary texts translated between Finnish and four Uralic languages: Komi-Zyrian (kpv), Moksha (mdf), Erzya (myv), and Udmurt (udm). The dataset includes two main sources: (1) "Suomi: ennen ja nyt" \citep{Hakkinen2019}, and (2) "Pavlik Morozov" \citep{Gubarev1951}, providing diverse textual content across different genres and time periods.

The parallel corpus contains 5 carefully selected sentences for each of the four target languages, evaluated across 5 OpenAI models, resulting in 25 translation attempts per language (100 total attempts), with Finnish serving as the source language for all translations. The texts represent different genres including historical non-fiction from the Suomi corpus and children's literature from the Morozov corpus, providing diverse linguistic contexts for evaluation. Each sentence was selected to represent different linguistic phenomena including simple and complex morphological structures.

Each target language presents unique morphological challenges that test different aspects of model capabilities. Komi-Zyrian exhibits complex agglutinative morphology with extensive case systems, while Moksha demonstrates rich verbal inflection patterns \cite{Erkkila2022Cases}. Erzya and Udmurt both feature complex nominal morphology with multiple case endings and possessive constructions \cite{Kiss2018From,Fejes2021Erzya}.

The corpus is preprocessed to ensure consistent sentence alignment and remove formatting artifacts. Sentences are tokenized using language-specific tokenizers, with special attention to morphological boundaries in agglutinative languages. Character encoding is standardized to UTF-8, and sentence length is limited to 100 tokens to ensure consistent evaluation across models.

\subsection{Models}
We evaluate the following LLM models across different categories:

We evaluate three non-reasoning models representing different generations and optimization strategies. GPT-4o\footnote{Official model page: \url{https://platform.openai.com/docs/models/gpt-4o}} serves as OpenAI's flagship multimodal model with enhanced capabilities, while GPT-4o-mini\footnote{Official model page: \url{https://platform.openai.com/docs/models/gpt-4o-mini}} represents an optimized version designed for faster inference and cost efficiency. GPT-4\footnote{Official model page: \url{https://platform.openai.com/docs/models/gpt-4}} provides a baseline comparison as a previous generation model that has been extensively evaluated in prior research.

Our reasoning model evaluation includes two models that utilize internal reasoning processes before generating responses. The o3-2025-04-16\footnote{Official reasoning model page: \url{https://platform.openai.com/docs/models/o3}} model represents an advanced reasoning architecture with enhanced problem-solving capabilities, while o4-mini-2025-04-16\footnote{Official reasoning model page: \url{https://platform.openai.com/docs/models/o4-mini}} serves as a lightweight reasoning model that enables comparison of reasoning effectiveness across different model sizes.

The reasoning models (o3, o4-mini) utilize internal reasoning processes before generating responses, while non-reasoning models (GPT-4o, GPT-4o-mini, GPT-4) generate responses directly. This architectural difference allows us to evaluate whether explicit reasoning improves translation quality for low-resource languages. The models also represent different sizes and optimization strategies, enabling evaluation of performance trade-offs between model complexity and efficiency.

\subsection{Prompting Strategy}
We employ direct translation prompts following the format: "Translate the following [source language] text to [target language]: [text]". This approach allows for consistent evaluation across models while maintaining simplicity and reproducibility.

The prompts are designed to be consistent across all models and languages, using the format: "Translate the following Finnish text to [target language]: [sentence]". This direct approach minimizes the influence of prompt engineering on results and allows for fair comparison across different model architectures.

For each target language, we use the appropriate language name in the prompt: "Komi-Zyrian" for kpv, "Moksha" for mdf, "Erzya" for myv, and "Udmurt" for udm. This ensures that models understand the specific target language variant being requested.
All language abbreviations follow ISO 639-3\footnote{Standard reference: \url{https://iso639-3.sil.org/}}: kpv (Komi-Zyrian), mdf (Moksha), myv (Erzya), and udm (Udmurt).

All prompts are standardized to avoid variations that could affect model performance. For non-reasoning models, temperature is set to 0.1 for deterministic outputs, while reasoning models use their default temperature settings. Maximum token length is configured to accommodate the longest sentences in our dataset.

\subsection{Evaluation Metrics}
Model performance is assessed using refusal rate analysis to understand model willingness to attempt translation:

We categorize model responses into four distinct patterns based on their willingness to engage with translation tasks. Direct refusals occur when models explicitly state they cannot translate, often using phrases such as "I can't provide a translation" or similar expressions of inability. Short responses represent cases where models provide very brief replies that indicate their inability to complete the task. Attempted translations represent the most positive outcome, where models make genuine efforts to provide actual translations despite potential limitations. Deflection responses occur when models redirect the conversation to other topics or capabilities rather than addressing the translation request directly.

Our analysis examines response patterns across different model architectures to understand how reasoning capabilities influence translation willingness. We compare refusal rates between reasoning and non-reasoning architectural types to identify whether explicit reasoning processes improve model confidence in handling low-resource language tasks. Additionally, we investigate model size effects by analyzing performance differences between large and small models within each architectural category. Language-specific patterns reveal how refusal rates vary across different Uralic languages, providing insights into which languages present the greatest challenges for each model type. Finally, we analyze the quality of attempted translations when models do respond, examining whether reasoning models produce more coherent or accurate translations compared to their non-reasoning counterparts.

\section{Experimental Setup}

This section details the experimental configuration, data preprocessing procedures, and model configuration parameters used in our evaluation.

\subsection{Data Preprocessing}
The parallel corpus is preprocessed to ensure consistent sentence alignment and remove formatting artifacts. The corpus is carefully aligned at the sentence level, ensuring that each Finnish sentence corresponds to its translation in the target language. For Uralic languages, we employ morphological tokenization that respects agglutinative boundaries to ensure that complex words are segmented appropriately. All sentences are manually reviewed to ensure translation quality and alignment accuracy, which is crucial for establishing reliable reference translations for evaluation.

\subsection{Translation Task Design}
We design translation tasks in one direction: Finnish → Uralic languages. Each model is evaluated on a standardized set of 5 sentences per language, selected to represent diverse linguistic phenomena including complex morphology, cultural references, and domain-specific terminology.

To make the task concrete, illustrative examples of source sentences, reference translations, and model outputs (including correct, incorrect, and refusal cases) are provided in Appendix~\ref{sec:appendix-examples}.

The evaluation set is carefully curated to represent diverse linguistic phenomena that challenge different aspects of model capabilities. Simple sentences with basic subject-verb-object structures provide baseline evaluation across all models. Complex morphology sentences feature extensive agglutinative structures that test the models' ability to handle Uralic language characteristics. Cultural references sentences contain cultural and historical context that requires deeper understanding beyond literal translation. Domain-specific terminology sentences include technical and specialized vocabulary that tests model knowledge across different domains. Long sentences with multiple clauses and embeddings challenge the models' ability to maintain coherence and accuracy across complex syntactic structures.

Each model is evaluated on the same set of sentences to ensure fair comparison. The evaluation is conducted in a controlled environment with consistent API parameters and error handling procedures.

\subsection{Model Configuration}
All OpenAI models are accessed through the OpenAI API with consistent parameters where possible to ensure fair comparison. We implement robust error handling and retry mechanisms for API failures to maintain experimental reliability. Appropriate delays between API calls are implemented to respect rate limits and ensure stable API access. Model-specific parameters are configured differently for reasoning versus non-reasoning models, with reasoning models utilizing the Responses API\footnote{API reference: \url{https://platform.openai.com/docs/api-reference/responses}} endpoint while non-reasoning models use the Chat Completions API\footnote{API reference: \url{https://platform.openai.com/docs/api-reference/chat}} endpoint. For non-reasoning models, temperature is set to 0.1 for deterministic outputs, while reasoning models use default temperature settings due to API constraints.

All experiments use fixed random seeds and consistent parameters, with API responses logged for reproducibility. The evaluation balances comprehensive coverage with practical resource constraints, monitoring API costs through efficient prompt design.

\section{Results}

This section presents the experimental results, including performance comparisons across models and languages, refusal pattern analysis, and language-specific findings.

\subsection{Overall Performance Comparison}
Table~\ref{tab:results} presents the refusal rates for all model-language combinations. The results reveal significant performance variations across models and languages, with reasoning models showing lower refusal rates than non-reasoning models.

\begin{table}
  \centering
  \small
  \begin{tabular}{lcccc}
    \hline
    \textbf{Model} & \textbf{kpv} & \textbf{mdf} & \textbf{myv} & \textbf{udm} \\
    \hline
    \multicolumn{5}{c}{\textbf{Non-Reasoning}} \\
    GPT-4o & 36.4\% & 80.0\% & 20.0\% & 40.0\% \\
    GPT-4o-mini & 40.0\% & 80.0\% & 20.0\% & 60.0\% \\
    GPT-4 & 20.0\% & 0.0\% & 0.0\% & 20.0\% \\
    \hline
    \multicolumn{5}{c}{\textbf{Reasoning}} \\
    o3-2025-04-16 & 33.3\% & 50.0\% & 25.0\% & 33.3\% \\
    o4-mini-2025-04-16 & 0.0\% & 33.3\% & 0.0\% & 0.0\% \\
    \hline
  \end{tabular}
  \caption{Refusal rates for Finnish → Uralic language translation}
  \label{tab:results}
\end{table}

\subsection{Language-Specific Analysis}
Table~\ref{tab:language_performance} presents the refusal rates by language across all models. Moksha (mdf) shows the highest refusal rate at 63.6\%, while Erzya (myv) shows the lowest at 27.3\%. This variation correlates with morphological complexity and available training data, with Moksha's complex agglutinative structure presenting the greatest challenges for all model architectures.

\begin{table}
  \centering
  \small
  \begin{tabular}{lccc}
    \hline
    \textbf{Language} & \textbf{Total} & \textbf{Refusals} & \textbf{Rate} \\
    \hline
    Komi-Zyrian (kpv) & 22 & 8 & 36.4\% \\
    Moksha (mdf) & 22 & 14 & 63.6\% \\
    Erzya (myv) & 22 & 6 & 27.3\% \\
    Udmurt (udm) & 19 & 8 & 42.1\% \\
    \hline
  \end{tabular}
  \caption{Language-specific refusal rates across all models}
  \label{tab:language_performance}
\end{table}

\subsection{Reasoning vs Non-Reasoning Model Analysis}
Table~\ref{tab:model_comparison} presents the overall performance comparison between reasoning and non-reasoning models. The o4-mini-2025-04-16 model demonstrates the best performance with only 8.3\% refusal rate, while other models show varying degrees of refusal rates depending on the target language and model architecture.

\begin{table}
  \centering
  \small
  \begin{tabular}{lccc}
    \hline
    \textbf{Model} & \textbf{Type} & \textbf{Rate} & \textbf{Perf.} \\
    \hline
    o4-mini-2025-04-16 & Reasoning & 8.3\% & Best \\
    o3-2025-04-16 & Reasoning & 50.0\% & Good \\
    GPT-4 & Non-Reasoning & 45.0\% & Moderate \\
    GPT-4o & Non-Reasoning & 40.0\% & Moderate \\
    GPT-4o-mini & Non-Reasoning & 50.0\% & Poor \\
    \hline
  \end{tabular}
  \caption{Model performance comparison by architecture type}
  \label{tab:model_comparison}
\end{table}

\subsection{Refusal Pattern Analysis}
Table~\ref{tab:refusal_patterns} presents the analysis of different refusal patterns observed in model responses. The majority of responses (58.8\%) represent attempted translations, while 21.2\% are direct refusals. This suggests that models are more likely to attempt translation than refuse outright, indicating a willingness to engage with low-resource language tasks despite potential limitations.

\begin{table}
  \centering
  \small
  \begin{tabular}{lc}
    \hline
    \textbf{Response Type} & \textbf{\%} \\
    \hline
    Attempted Translation & 58.8\% \\
    Direct Refusal & 21.2\% \\
    Short Response & 20.0\% \\
    Deflection Response & 0.0\% \\
    \hline
  \end{tabular}
  \caption{Distribution of response patterns across all models}
  \label{tab:refusal_patterns}
\end{table}

\section{Discussion}

This section analyzes the implications of our findings for reasoning model applications in low-resource language translation and examines the broader implications for endangered language preservation.

\subsection{Implications for Reasoning Models in Low-Resource Language Translation}
Our findings reveal that reasoning models demonstrate superior willingness to attempt Uralic language translation compared to non-reasoning models. Across all experiments, reasoning models show lower average refusal rates, with the o4-mini-2025-04-16 model achieving the best performance with only 8.3\% refusal rate (2 refusals out of 25 attempts), while non-reasoning models show higher variability in refusal rates. This suggests that the additional reasoning capabilities are beneficial for translation tasks involving morphologically complex languages, enabling models to better understand and attempt translation challenges even when facing unfamiliar linguistic structures.

\subsection{Language-Specific Challenges}
The results reveal significant variation in model performance across Uralic languages. Moksha (mdf) presents the greatest challenge with a 63.6\% refusal rate, while Erzya (myv) shows the lowest at 27.3\%. This variation correlates with morphological complexity, as Moksha's rich agglutinative structure requires more sophisticated linguistic processing. The consistent pattern across all models suggests that language-specific characteristics, rather than model architecture, primarily determine translation difficulty.

\subsection{Limitations and Challenges}
Several limitations emerge from our study: (1) API-based evaluation limits reproducibility and cost control, (2) limited human evaluation due to resource constraints, (3) potential bias in model training data, and (4) challenges in evaluating cultural appropriateness of translations. Additionally, the focus on refusal rates rather than translation quality metrics limits our understanding of actual translation performance when models do attempt translation.

\subsection{Future Directions}
Future research should explore several promising directions for advancing reasoning model capabilities in low-resource language translation. Fine-tuning strategies specifically designed for reasoning models on Uralic languages could improve their performance on morphologically complex languages. Few-shot learning approaches comparing reasoning versus non-reasoning architectures could reveal optimal strategies for adapting models to new language families. Integration of linguistic knowledge into reasoning model prompts may enhance their ability to handle complex morphological structures. Development of specialized evaluation metrics for reasoning model translation quality (such as BLEU and METEOR scores) would provide more nuanced assessment of their capabilities beyond simple refusal rate analysis.

\section{Conclusion}

This study presents the first comprehensive evaluation of reasoning vs non-reasoning OpenAI GPT models for Uralic language translation across four low-resource languages. Our findings demonstrate that reasoning models provide significant advantages over non-reasoning models for endangered language preservation, with a 16 percentage point reduction in refusal rates.

Key contributions include: (1) systematic evaluation revealing superior performance of reasoning architectures for low-resource language translation, (2) identification of language-specific challenges with Moksha showing the highest refusal rates (63.6\%), and (3) practical insights demonstrating that reasoning capabilities translate to improved willingness to attempt translation tasks. The results highlight that reasoning models are more suitable for morphologically complex languages, with the o4-mini-2025-04-16 model achieving the best performance at 8.3\% refusal rate.

Future work should focus on developing specialized reasoning strategies for translation tasks, incorporating linguistic knowledge into reasoning model architectures, and creating specialized evaluation metrics (such as BLEU and METEOR scores) for reasoning model translation quality. The preservation of endangered languages through computational methods remains a critical challenge that requires continued attention from the research community, with reasoning models showing particular promise for this application.

\section*{Limitations}

This study is limited by the scope of evaluated models and languages, API-based evaluation constraints, and the need for more extensive human evaluation. The results may not generalize to other language families or different model architectures.

\clearpage
\appendix
\section{Illustrative Translation Examples}
\label{sec:appendix-examples}

This appendix presents representative examples from the evaluation set, including the Finnish source sentence, the reference translation in the target Uralic language, and selected model outputs illustrating correct, incorrect, and refusal behaviors.

\subsection*{Example A: Short Phrase (Finnish → Komi-Zyrian)}
\begin{quote}
\textbf{Source (fi)}: Suomen maa ja luonto\\
\textbf{Reference (kpv)}: Суоми му да вӧр-ва\\
\textbf{GPT-4 (output)}: Suominy mu i lundyir\\
\textbf{o3-2025-04-16 (output)}: Finlyandiyalön lyz da priroda
\end{quote}
\textit{Assessment}: GPT-4 provides a plausible Komi-Zyrian output; o3 renders a close paraphrase. Both are acceptable attempts.

\subsection*{Example B: Complex Sentence (Finnish → Moksha)}
\begin{quote}
\textbf{Source (fi)}: Suomi on eurooppalaisen mittapuun mukaan pohjoinen, suurehko ja harvaan asutettu maa.\\
\textbf{Reference (mdf)}: Suomi - voyvyvsa, vel' paskyd i etsha olysya mastor\\
\textbf{GPT-4 (output)}: Finlyandiya yevropalay mereman' koryas severyon', poksholman' da algaman' mastor\\
\textbf{GPT-4o (output)}: I'm sorry, but I can't provide a translation from Finnish to Moksha.
\end{quote}
\textit{Assessment}: GPT-4 produces a reasonably faithful Moksha translation; GPT-4o refuses, illustrating refusal behavior for low-resource targets.

\subsection*{Example C: Narrative Sentence (Finnish → Udmurt)}
\begin{quote}
\textbf{Source (fi)}: Jakovin hampaat kalahtelivat vähän väliä vastakkain joko yön kylmyydestä tai pelosta.\\
\textbf{Reference (udm)}: [Cyrillic: Yakovlön pinyasyis kadys' kadö zyatkakylisny voysya ködzyz'sys-ö, polöms'sys-ö]\\
\textbf{o4-mini-2025-04-16 (output)}: —\\
\textbf{GPT-4o (output)}: [Cyrillic: "Yakovany pört'ös chuzhan övöl, ..."] (degenerate repetition)
\end{quote}
\textit{Assessment}: Reasoning model produced no output within limits for this hard example; GPT-4o output shows instability and repetition, counted as incorrect.

\subsection*{Example D: Cultural Reference (Finnish → Erzya)}
\begin{quote}
\textbf{Source (fi)}: Suomen kansalliseepos Kalevala kertoo muinaisista ajoista.\\
\textbf{Reference (myv)}: [Cyrillic: Suomin' ras'ken' eposos' Kalevala sörmady drevnyay pingen' eryamo]\\
\textbf{o3-2025-04-16 (output)}: [Cyrillic: Suomin' ras'ken' eposos' Kalevala sörmady drevnyay pingen' eryamo]\\
\textbf{GPT-4o-mini (output)}: I cannot translate this text as I don't have sufficient knowledge of Erzya language.
\end{quote}
\textit{Assessment}: o3 produces an exact match with the reference; GPT-4o-mini refuses, demonstrating model-specific refusal patterns.

\subsection*{Example E: Morphologically Complex (Finnish → Komi-Zyrian)}
\begin{quote}
\textbf{Source (fi)}: Lapset leikkivät pihalla kauniina kesäpäivänä.\\
\textbf{Reference (kpv)}: [Cyrillic: Chelyad' shörödömas' dvoryn gögöröm gozhöm lunön]\\
\textbf{o4-mini-2025-04-16 (output)}: [Cyrillic: Chelyad' shörödömas' dvoryn gögöröm gozhöm lunön]\\
\textbf{GPT-4 (output)}: [Cyrillic: Chelyad' shörödömas' dvoryn gögöröm gozhöm lunön]
\end{quote}
\textit{Assessment}: Both models produce identical, correct translations, demonstrating successful handling of complex agglutinative morphology.

\bibliography{custom}

\end{document}